\documentclass{article}
\usepackage{times}
\usepackage{helvet}
\usepackage{courier}
\usepackage{epsfig}
\usepackage{natbib}
\usepackage{hyperref}
\usepackage{algorithm}
\usepackage{algorithmic}
\usepackage{caption}
\usepackage{subcaption}
\usepackage{mathrsfs}
\usepackage{tikz}
\usetikzlibrary{automata,chains,arrows}
\usetikzlibrary{positioning} 
\tikzset{
    >=stealth',
    punkt/.style={
           rectangle,
           rounded corners,
           draw=black, very thick,
           text width=6.5em,
           minimum height=2em,
           text centered},
    pil/.style={
           ->,
           thick,
           shorten <=2pt,
           shorten >=2pt,}
}
\usepackage{color}
\usepackage{pifont} 
\usepackage{amsmath}
\usepackage{amsthm}
\usepackage{amssymb}
\usepackage{colortbl}

\newtheorem{theorem}{Theorem}
\newtheorem{lemma}{Lemma}
\newtheorem{definition}{Definition}
\newtheorem{example}{Example}

\newtheorem{remark}{Remark}

\DeclareMathOperator*{\argmax}{arg\,max}

\begin{document}
\title{Online Transfer Learning in Reinforcement Learning Domains}
\author{Yusen Zhan, Matthew E. Taylor  \\
	School of Electrical Engineering and Computer Science\\
	Washington State University \\
    \{yzhan,taylorm\}@eecs.wsu.edu}
\maketitle

\begin{abstract}
\begin{quote}
This paper proposes an online transfer framework to capture the interaction among agents and shows that current transfer learning in reinforcement learning is a special case of online transfer. Furthermore, this paper re-characterizes existing agents-teaching-agents methods as online transfer and analyze one such teaching method in three ways. First, the convergence of Q-learning and Sarsa with tabular representation with a finite budget is proven. Second, the convergence of Q-learning and Sarsa with linear function approximation is established. Third, the we show the asymptotic performance cannot be hurt through teaching. Additionally, all theoretical results are empirically validated.
\end{quote}
\end{abstract}

\section{Introduction}
Agents can autonomously learn to master sequential decision tasks by reinforcement learning~\cite{sutton1998introduction}. Traditionally, reinforcement learning agents are trained and used in isolation. More recently, the reinforcement learning community became interested in interaction among agents to improve learning.

%removed reference to humans
%, or between agents and humans.

There are many possible methods to assist agent's learning~\cite{erez2008does,taylor2009transfer}. This paper focuses on \emph{action advice}~\cite{torrey2013teaching}: as the student agent practices, the teacher agent suggests actions to take. This method requires only agreement of the action sets between teachers and students, while allowing for different state representations and different learning algorithms among teachers and students. 

Although this advice method is shown to empirically provides multiple benefits~\cite{torrey2013teaching,LIP610219}, existing work does not provide a formal understanding of teaching or advice. Therefore, this paper proposes a framework --- an online transfer framework --- to characterize the interaction among agents, aiming to understand the teaching or advice from the transfer learning perspective. We extend the transfer learning framework in reinforcement learning proposed by Lazaric (\citeyear{transferalessan}) %\MET{We don't want to use citations as nouns. You can use the author's name as a noun, and put the year in parentheses} 
into online transfer learning which capture the online interaction between agents. Also, we show that 1) transfer learning is a special case of online transfer framework, and 2)
%Active learning is a branch of machine learning which is related supervised learning \cite{settles2010active}. 
%The key idea of active learning is that there is a oracle provides labels to learning algorithms, speeding up the learning process.  
our framework is similar to that of of active learning~\cite{settles2010active}, but in a reinforcement learning setting.

After introducing our novel framework, it can be used to analyze existing advice methods, such as action advice \cite{torrey2013teaching}. 
First, we prove the convergence of Q-learning and Sarsa with tabular representation with a finite amount of advice. Second, the convergence of Sarsa and Q-learning with linear function approximation is established with finite advice. The convergence means the algorithms converge to the optimal Q-value. Third, we show that a non-infinite amount of advice cannot change the student's asymptotic performance. These three results are then confirmed empirically in a simple Linear Chain MDP and a more complex Pac-Man simulation.

%\MET{We should say "First... Second... Third..."  Not "First... Second... Finally"}

\section{Background}
This section provides necessary background, adopting some notation introduced elsewhere \cite{sutton1998introduction,melo2008analysis}.
%In this section, we provide necessary background for this paper. %Firstly, we introduce the basic notation about the Markov Decision Process and related estimation definitions such as $V$, $V^\star$, $Q$ and $Q^\star$. Then, we review some reinforcement learning algorithms---Q-learning and SARSA and corresponding linear function approximation. Finally, we formally define advice model in our previous work and explain it in online TR framework. 
%We adopt some notations from \cite{melo2008analysis,sutton1998introduction}.
%\newpage
\subsection{Markov Decision Process}
Let $M= \langle S,A,P,R,\gamma \rangle $ be a Markov decision process (MDP) with a compact state set $S$ and a finite action set $A$. $P$ is the transition probability kernel. For any $(s,a,s')\in S \times A\times S$ triplet the probability of transition from state $s$ taking action $a$ to state $s'$ is defined as 
         $\mathcal{P}[s'\in U| s,a]=P(U|s,a),$
where $U$ is a Borel-measurable subset\footnote{Details on Borel-measurable subsets can be found elsewhere~\cite{rudin1986real}.} of $S$. $R: S \times A \times S\to \mathbb{R}$ is a bounded deterministic function which assigns a reward $R(s,a,s')$ to transition from state $s$ to state $s'$ taking action $a$. The discount factor is $\gamma$ such that $0 \leq \gamma \leq 1$.  
The expected total discounted reward for $M$ under some policy can be defined as $\mathbb{E}\left[\sum_{t=0}^{\infty}\gamma^t r(s_t,a_t) \right],$
where $t$ is the time step and $r(s_t,a_t)$ denotes the reward received for taking action $a_t$ in state $s_t$ at time step $t$, according to reward distribution. For convenience, we omit the state and the action and only use $r_t$ to denote the reward received at time step $t$, so the expected total discounted reward can be written as $\mathbb{E}\left[\sum_{t=0}^{\infty}\gamma^t r_t \right]$. $r(s,a)$ and $R(s,a,s')$ have following relationship: $\mathbb{E}[r(x,a)]=\int_S R(s,a,s')P(ds'|s,a)$. %$\text{\it{Var}}(r(x,a))=\int_S (\mathbb{E}[r(x,a)]-R(x,a,y))^2P(dy|x,a)$.

A policy is a mapping that outputs for each state-action pair $(s,a)$. A deterministic policy $\pi$ is a mapping defined as $\pi : S \to A$, while a stochastic policy is a mapping defined over $S \times A$ (i.e., $\mathcal{P}[\text{choose action } a| \text{at state } s]=\pi(s,a).$)

The state-action function
%,  $Q^{\pi}(s,a)$, of taking action $a$ in state $s$ under a policy $\pi$ can be written as:
is the expected return for a state action pair under a given policy:
$Q^{\pi}(s,a)=\mathbb{E}_{\pi} \left[ \sum_{k=0} ^{\infty}\gamma^{k} r_{t+k} \middle| s_t=s,a_t=a\right]$.
Solving an MDP 
%$M=(S,A,P,R,\gamma)$ 
usually means finding an optimal policy that maximizes the expected return. An optimal policy $\pi^\star$ is such that $Q^{\pi^\star}\geq Q^{\pi}$ for all $s \in S$, all $a\in A$ and all policies $\pi$. We can define the optimal state-value function $Q^\star$ as 
$Q^\star(s,a) =\int_S (R(s,a,s')+\gamma\max_{a'\in A}Q^\star (s',a'))P(ds'|s,a)$,
representing the expected total discounted reward received along an optimal trajectory when taking action $a$ in state $s$  and following optimal policy $\pi^\star$ thereafter.
For all $s \in S$,
$\pi^\star(s)=\argmax_{a \in A} Q^\star(s,a).$
%Notice that the optimal $Q^\star$ can be obtained from an optimal deterministic policy $\pi^\star$, even though stochastic policies may also be optimal.
Notice that although a stochastic policy may be optimal, there will always be a deterministic optimal policy with at least as high an expected value.

\subsection{Q-learning and Sarsa}
Q-learning is an important learning algorithm in reinforcement learning. It is a model-free and off-policy learning algorithm which is a break-through in reinforcement learning control. \citeauthor{watkins1989learning} (\citeyear{watkins1989learning}) introduced Q-learning as follows: 

Given any estimate $Q_0$, Q-learning algorithm can be represented by following update rules:
\begin{equation}
\label{q-learning}
Q_{t+1}(s,a)  =Q_t(s,a)+\alpha_t(s,a)\Delta_t
\end{equation}
where $Q_t$ denotes the estimation of $Q^\star$ at time $t$, $\{\alpha_t(s,a)\}$ denotes the step-size sequence and $\Delta_t$ denotes temporal difference at time $t$,
\begin{equation}
\label{q-update}
\Delta_t=r_t +\gamma\max_{a' \in A} Q_t(s',a')-Q_t(s,a)
\end{equation}
where $r_t$ is the reward received at time step $t$. The update Equation \ref{q-update} does not dependent any policies, so the Q-learning is called off-policy algorithm.

%
%In contrast to off-policy algorithms, there are some on-policy algorithms in which Sarsa is the analogy of Q-learning \cite{rummery1994line}. Given any estimate $Q_0$ and a learning policy $\pi$, SarsaR algorithm can be represented by following update rules:
%\begin{equation}
%\label{Sarsa}
%Q_{t+1}(s,a)  =Q_t(s,a)+\alpha_t(s,a)\Delta_t
%\end{equation}
%where $\{\alpha_t(s,a)\}$ denotes the step-size sequence and $\Delta_t$ denotes temporal difference at time $t$,
%\begin{equation}
%\label{sarsa-update}
%\Delta_t=r_t +\gamma Q_t(s',a')-Q_t(s,a)
%\end{equation}
%where $a'$ is determined by the policy $\pi$ and $r_t$ is the reward received at time step $t$. Notice that the Equation \ref{sarsa-update} involves with policy $\pi$, therefore, it is on-policy.

In contrast to off-policy algorithms, there are some on-policy algorithms in which Sarsa is the analogy of Q-learning \cite{rummery1994line}. Given any estimate $Q_0$ and a policy $\pi$, the difference between Q-learning and Sarsa is that the temporal difference $\Delta_t$: 
\begin{equation}
\label{sarsa-update}
\Delta_t=r_t +\gamma Q_t(s',a')-Q_t(s,a)
\end{equation}
where $a'$ is determined by the policy $\pi$ and $r_t$ is the reward received at time step $t$. Notice that the action selection in Equation \ref{sarsa-update} involves the policy $\pi$, making it on-policy.

If both $S$ and $A$ are finite sets, the Q-value function can be easily represented by an $|S| \times |A|$ matrix and it can be represented in a computer by a table. This matrix representation is also called tabular representation. In this case, the convergence of Q-learning, Sarsa, and other related algorithms (such as TD($\lambda$)) have been shown by previous work \cite{peter1992convergence,watkins1992q,singh2000convergence}. %\MET{Order citations by year, and then alphabetically by 1st author last name}
However, if $S$ or $A$ is infinite or very large, it is infeasible to use tabular representation and a compact representation is required (i.e., function approximation). %\MET{when you use e.g. or i.e., use a comma after the abbreviation} 
This paper focuses on Q-learning with linear function approximation and Sarsa with linear function approximation. The linear approximation means that state-value function $Q$ can be represented by a linear combination of features $\{\phi_i\}_{i=1}^{d}$, where $\phi_i: S\times A \to \mathbb{R}$ is the feature and $d$ is the number of features. Given a state $s\in S$ and an action $a\in A$, the action value at time step $t$ is defined as 
\begin{equation}
Q_t(s,a)=\sum_{i=1}^d \theta_t(i)\phi_i(s,a)=\pmb\theta^T_t\pmb\phi(s,a)
\end{equation}
where $\pmb\theta_t$ and $\pmb\phi$ are $d$-dimensional column vectors and $^T$ denotes the transpose operator. Since $\pmb\phi$ is fixed, algorithms only are able to update $\pmb\theta_t$ each time. Gradient-descent methods are one of most widely used of all function approximation methods. Applying a gradient-descent method to Equation \ref{q-learning}, we obtain approximate Q-learning:
\begin{align}
\label{q-learninglinearapproximation}
\begin{split}
 \pmb\theta_{t+1} &= \pmb\theta_t+\alpha_t(s,a)\nabla Q_t(s,a)\Delta_t \\
  &=\pmb\theta_t+\alpha_t(s,a)\pmb\phi(s,a)\Delta_t
\end{split}
\end{align}
where $\{\alpha_t\}$ %\MET{Why put alpha-t inside brackets?}
 is the update parameter at time $t$ and $\Delta_t$ is the temporal difference at time step $t$ (Equation \ref{q-update}). Similarly, given a policy $\pi$, the on-policy temporal difference can be defined as
\begin{align}
\label{Sarsalinearupdate}
\begin{split}
\Delta_t&=r_t +\gamma Q_t(s',a')-Q_t(s,a) \\
&=r_t+\gamma\pmb\theta^T_t\pmb\phi(s',a')-\pmb\theta^T_t\pmb\phi(s,a),
\end{split}
\end{align}
where $a'$ is determined by the policy $\pi$ at time $t$. Combining Equation \ref{q-learninglinearapproximation} and Equation \ref{Sarsalinearupdate}, we obtain Sarsa with linear approximation. For a set fixed features $\{\Phi_i:S\times A \to \mathbb{R}\}$, our goal is to learn a parameter vector $\pmb\theta_{\star}$ such that $\pmb\theta_{\star}^\intercal\pmb\Phi(s,a)$ approximates the optimal Q-value $Q^\star$.

\section{Online Transfer Framework}
%In this section, we proposed a framework, which is inspired by the framework introduced by Lazaric for transfer learning in reinforcement learning paradigm, for online transfer learning in reinforcement learning domain at first \cite{transferalessan}. Then, we compare our framework to Lazaric's work on transfer learning.
This section introduces a framework for online transfer learning in reinforcement learning domains, inspired by previous work \cite{transferalessan}.
%\MET{don't use contractions in papers.}

\subsection{Online Transfer}

Transfer learning is a technique that leverages past knowledge in one or more \emph{source tasks} to improve the learning performance of a learning algorithm in a \emph{target task}. Therefore, the key is to describe the knowledge transferred between different algorithms. A standard reinforcement learning algorithm usually takes input some raw knowledge of the task and returns a solution in a possible set of solutions. We use $\mathscr{K}$ to denote the space of the possible input knowledge for learning algorithms and $\mathscr{H}$ to denote the space of hypotheses (possible solutions, e.g., policies and value functions). Specifically, $\mathscr{K}$ refers to all the necessary input information for computing a solution of a task, e.g., samples, features and learning rate. 

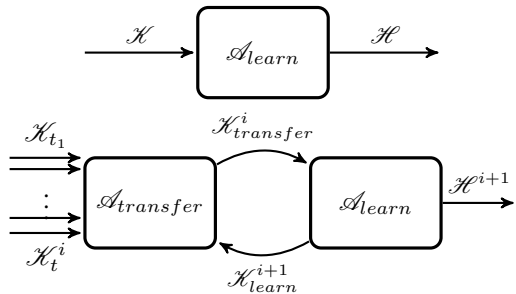
\begin{figure}[t]
\centering
\begin{tikzpicture}[->,>=stealth',shorten >=1pt,auto,node distance=0.5cm,
  thick,main node/.style={circle,fill=blue!20,draw,font=\sffamily\Large\bfseries}]
 %nodes
 \node[punkt,minimum height=1.2cm, text width=1.5cm] (market) at (3.5,0) {$\mathscr{A}_{learn}$};
 \node(dummy1) at (1,0) {};
 \node(dummy2) at (6,0) {};
  
 %\node[below=1cm of market](dummy3) {};
 \node[punkt,minimum height=1.2cm, text width=1.5cm] (m1) at (2,-2){$\mathscr{A}_{transfer}$};
 \node[punkt,minimum height=1.2cm, text width=1.5cm] (m2) at (5,-2){$\mathscr{A}_{learn}$};
 \node (dummy4) at (0,-2) {};
 \node (dummy5) at (7,-2) {};
  
 \path[every node/.style={font=\sffamily\small}]
  
  (dummy1) edge node [above] {$\mathscr{K}$}  (market)
  
  (market) edge node [above] {$\mathscr{H}$} (dummy2)
  
  (m1) edge[bend left] node [above] {$\mathscr{K}_{transfer}^{i}$}  (m2)
  
  (m2) edge node[above] {$\mathscr{H}^{i+1}$} (dummy5)
  
  		  edge[bend left] node[below]  {$\mathscr{K}_{learn}^{i+1}$} (m1);
\path  (dummy4) edge [transform canvas={yshift=6mm}] node[above] {$\mathscr{K}_{t_1}$} (m1);
\path  (dummy4) edge [transform canvas={yshift=4.5mm}] node[below] {$\vdots$} (m1);

\path  (dummy4) edge [transform canvas={yshift=-2mm}] (m1);
\path  (dummy4) edge [transform canvas={yshift=-4mm}] node[below] {$\mathscr{K}_t^{i}$} (m1);
\end{tikzpicture}
\caption{\textit{(Top)} The standard learning process only requires original knowledge from the target tasks. \textit{(Bottom)} In the online transfer learning process, transfer algorithm takes input knowledge from source tasks, target task and learner at time $i$ and output transfer knowledge at time $i$, then the leaning algorithm takes the transfer knowledge at time $i$ to generate hypothesis at time $i+1$. This process will repeat until a good hypothesis is computed.}
\label{onlinetransferframework}
\end{figure}

In general, the objective of transfer learning is to reduce the need for samples from the target task by taking advantage of prior knowledge. An online transfer learning algorithm can been defined by a sequence of transferring and learning phases, e.g., 1) transferring knowledeg, 2) learning, 3) transferring based on previous learning, 4) learning, etc. Let $\mathscr{K}^{L}_{s}$ be the knowledge from $L$ source tasks, $\mathscr{K}_{t}^{i}$ be the knowledge collected from the target task at time $i$ and $\mathscr{K}_{learn}^{i}$ be the knowledge obtained from learning algorithm at time $i$ (including previous learning phases). We define one time step as one-step update in a learning algorithms or one batch update in batch learning algorithms. Thus, the algorithm may transfer \textbf{one-step} knowledge, or \textbf{one-episode} knowledge, or even \textbf{one-task} or \textbf{multi-task} knowledge to the learner, depending on the setting. $\mathscr{K}^{i}$ denote the knowledge space with respect to time $i$ such that $\mathscr{K}^{i} \subseteq \mathscr{K}$, for all $i=0,1,2,\dots$. The online transfer learning algorithm can be defined as 
\begin{equation}
\label{onlineTR}
\mathscr{A}_{transfer}:\mathscr{K}^{L}_{s} \times \mathscr{K}_{t}^{i}\times \mathscr{K}_{learn}^{i} \to \mathscr{K}_{transfer}^{i}
\end{equation}
where $\mathscr{K}_{transfer}^{i}$ denotes the knowledge transferred to the learning phase at time $i$, $i=0,1,2,\dots$. Notice that $ \mathscr{K}_{learn}^{i}$ is generated by the learning algorithm. Thus, the reinforcement learning algorithm can be formally described as 

\begin{equation}
\label{onlineLearning}
\mathscr{A}_{learn}:\mathscr{K}_{transfer}^{i} \times \mathscr{K}_{t}^{i} \to \mathscr{K}_{learn}^{i+1} \times \mathscr{H}^{i+1}
\end{equation}
where $\mathscr{K}_{learn}^{i+1}$ is the knowledge from learning algorithm at time $i+1$ and $\mathscr{H}^{i+1}$ is the hypothesis space at time $i+1$, $i=0,1,2,\dots$. $\mathscr{K}_{learn}^{i+1}$ is used as input for next time step in online transfer Equation \ref{onlineTR}. Then,  $\mathscr{A}_{transfer}$ generates the transferred knowledge $\mathscr{K}_{transfer}^{i}$ for learning phase in Equation \ref{onlineLearning}. $\mathscr{A}_{learn}$ computes the $\mathscr{K}_{learn}^{i+1}$ for the next time step, and so on. In practice, the initial knowledge from the learning phase, $\mathscr{K}_{learn}^{0}$ can be empty or any default value. In this framework, we expect the hypothesis space sequence $\mathscr{H}^{0}, \mathscr{H}^{1}, \mathscr{H}^2, \dots$ will become better and better over time under some criteria (e.g., the maximum average reward or the maximum discounted reward), where $\mathscr{H}^{i} \subseteq \mathscr{H}$ is the space of hypothesis with respect to $i$, $i=0,1,2,\dots$, that is, the space of possible solutions at time $i$. See Figure \ref{onlinetransferframework} for an illustration.

%\MET{Have we introduced ``Mistake Correcting'' at this point?}
%\MET{We shouldn't cite your workshop paper in this. We can put it into the final version, but the initial submission needs to be blind, and it's likely that this recent workshop paper is only known by us.}
%\MET{If we wanted, somewhere we could cite the following paper: \url{https://matthieu-zimmer.net/publications/ARMS2014.pdf}}
\begin{example}
Consider the Active Relocation Model \cite{mihalkova2006using}. In this setting, there is an expert and a learner, which can be treated as the transfer algorithm $\mathscr{A}_{transfer}$ and the learning algorithm $\mathscr{A}_{learn}$, respectively. The learner is able to relocate its current state to a visited state, but the learner may become stuck in a sub-optimal state. Thus, the expert is able to help the learner to relocate its current state to a better state according to the expert's knowledge. 
This algorithm can be represented in our framework as $\mathscr{K}_s=(S\times A \times S \times R)^{N_s}$, where $N_s$ is the number of samples the expert collect from the source tasks, $\mathscr{K}_t^{i}=(S_i\times A_i \times S_i \times R_i)^{N_i}$, $\mathscr{K}_{learn}^{i}=(\hat{Q}_i \times S_i \times A_i) $, $\mathscr{K}_{transfer}^{i}=(S_i)$ and $\mathscr{H}^{i+1}=\{\hat{Q}_{i+1}\}$, $i=0,1,\dots,n$\footnote{$S_i$, $A_i$, $R_i$ are all subsets of the set $S$ of states, the set $A$ of actions, and the set $R$ of rewards, respectively. $\hat{Q}_i$ is the estimate of $Q$-value function at time $i$. We introduce the index to distinguish the the difference in different time steps. For example, the learning algorithm is able to reach more states at time step $i+1$ than at time step $i$. Thus, $S_{i}\subseteq S_{i+1}$.}.
\end{example}

Although we explicitly introduce $\mathscr{K}_{t}^{i}$ and $\mathscr{K}_{learn}^{i}$ in Equation \ref{onlineTR} and \ref{onlineLearning}, in most settings, it is impossible for the transfer algorithm and learning algorithm 
to explicitly access the knowledge from target tasks or it only has a limited access to it. For example, the communication failure and restrictions may cause these problems.

\subsection{Transfer Learning and Online Transfer Learning}
Our online transfer learning framework can be treated as an online extension of 
%\MET{an existing (offline)} 
the transfer learning framework \cite{transferalessan}. If we set all  $\mathscr{K}_{learn}^{i}=\emptyset$ and set $i=0$, we have 
\begin{equation}
\label{TR}
\mathscr{A}_{transfer}:\mathscr{K}^{L}_{s} \times \mathscr{K}_{t}^{0}\times \emptyset \to \mathscr{K}_{transfer}^{0}
\end{equation}
\begin{equation}
\label{Learning}
\mathscr{A}_{learn}:\mathscr{K}_{transfer}^{L} \times \mathscr{K}_{t}^{0} \to \emptyset\times \mathscr{H}^{1}
\end{equation}
where the the $\mathscr{A}_{transfer}$ transfers the knowledge to the $\mathscr{A}_{learn}$ once, returning to the classic transfer learning scenario.

\subsection{Advice Model with Budget}
Now we discuss an advice method in previous work \cite{torrey2013teaching}, a concrete implementation of online transfer learning. Suppose that the teacher has learned an effective policy $\pi_t$ for a given task. Using this fixed policy, it will teach students beginning to learn the same task. As the student learns, the teacher will observe each state $s$ the student encounters and each action $a$ the student takes. Having a budget of $B$ advice, the teacher can choose to advise the student in $n\leq B$ of these states to take the ``correct'' action $\pi_t(s)$. 

The authors~\cite{torrey2013teaching} assumed the teacher's action advice is always correct and that students were required to execute suggested actions. Suppose that a reinforcement learning teacher agent $T$ is trained in a task and has access to its learned Q-Value function $Q_{t}$. Then, a student agent $S$ begins training in the task and is able to accept advice in the form of actions from the teacher. 
%MET removed the next sentence because we say it in the previous paragraph
%We further assume that the teacher has a limitation on the amount of advice he can give. 
We use notation $(T,S,\pi_d)$ to denote the advice model where $T$ is the teacher agent, $S$ is the student agent and $\pi_d$ is the policy teacher that provides its advice to student. The following example illustrates the how to characterize this advice model in the context of our online transfer learning framework.

\begin{example}
Let us consider the advice model using Mistake Correcting approach with limited budget and linear function approximation \cite{torrey2013teaching}. In this model, there is a teacher and a student, which can be treated as the transfer algorithm and the learning algorithm, respectively. First, the transfer algorithm $\mathscr{A}_{transfer}$ collects $N_s$ samples from $L$ source tasks. Then, it will return an advice action $a$ to the learning algorithm $\mathscr{A}_{learn}$ according to the current state and the action observed from the learning algorithm (initial knowledge is empty). The learning algorithm $\mathscr{A}_{learn}$ takes the advice action $a$ and $N_i$ samples from target task and returns a state and a action for next step, meanwhile, the $\mathscr{A}_{learn}$ maintains a function in the space $\mathscr{H}^{i+1}$ spanned by the features $\{\phi_j\}_{j=1}^{n}$, where $\phi_j: S \times A \to \mathbb{R}$ is defined by a domain expert. Moreover, the teacher has a limited budget $n$ for advising the student, so the time step $i=0,1,\dots,n$. Therefore, we have $\mathscr{K}_s=(S\times A \times S \times R)^{N_s}$, $\mathscr{K}_t^{i}=(S_i\times A_i \times S_i \times R_i)^{N_i}$, $\mathscr{K}_{learn}^{i}=(S_i \times A_i) $, $\mathscr{K}_{transfer}^{i}=(A_i)$ and $\mathscr{H}^{i+1}=\{f(\cdot,\cdot)=\sum_{j=1}^d\theta_{i+1}(j)\phi_j\}$, $i=0,1,\dots,n$.
\end{example}

\section{Theoretical Analysis}
For an advice model $(T,S,\pi_d)$ we propose in this paper, the most important theoretical problem is to resolve the convergence of algorithms\textbf{ since it guarantees the correctness of algorithms}. %Although we do not have convergence results for general methods such as $Q(\lambda)$ or $Sarsa(\lambda)$, the convergence for Q(0) (Q-learning), Q(0) with linear function approximation, Sarsa(0) and Sarsa(0) with linear function approximation has been investigated.   
In the next subsection, we will discuss how action advice interacts with the tabular versions of Q-learning and Sarsa. After, the corresponding algorithms with linear function approximation are discussed. 

\subsection{Tabular Representation}

The convergence of Q(0) (Q-learning) has been established by many works \cite{watkins1992q,jaakkola1994convergence,tsitsiklis1994asynchronous,mohri2012foundations}. 
%We will use the one of convergence results from \cite{mohri2012foundations}

\begin{lemma}(\cite{mohri2012foundations} Theorem 14.9 page 332)
\label{q-learninglemma}
Let $M$ be a finite MDP. Suppose that for all $s\in S$ and $a \in A$,  the step-size sequence $\{\alpha_t(s_t,a_t)\}$ such that 
\begin{equation*}
\sum_t \alpha_t(s_t,a_t)=\infty \qquad \sum_t \alpha_t(s_t,a_t)^2<\infty,
\end{equation*}
Then, the Q-learning Algorithm converges with probability 1.
\end{lemma}
Notice that the conditions on $\alpha_(s_t,a_t)$ ensure the infinity visits of action-state pairs.

\begin{theorem}
Given an advice model $(T,S,\pi_d)$, the student $S$ adopts the Q-learning Algorithm and conditions in Lemma \ref{q-learninglemma} all hold, convergence of Q-learning still holds in the advice model setting.
\end{theorem}

\begin{proof}
Notice that the conditions on $\alpha_t(s_t,a_t)$ verifies that each state-action pair is visited infinitely many times. And there is finite advice in our advice model. Therefore, the assumptions still hold in advice model setting. Apply Lemma \ref{q-learninglemma}, the convergence result follows. 
\end{proof}

Compared to Q-learning, Sarsa is a on-policy algorithm which requires a learning policy to update the Q values. \cite{singh2000convergence} prove that Sarsa with GLIE policy converges. We use their result to prove the convergence of Sarsa in advice model. First of all, we need to define GLIE policy.

\begin{definition}
A decaying policy $\pi$ is called GLIE, greedy in the limit with infinite exploration, policy, if it satisfies following two conditions:
\begin{itemize}
\item each state-action pair is visited infinity many times;
\item the policy is greedy with respect to the Q-value function with probability 1.
\end{itemize}
\end{definition}

It is not hard to verify that the Boltzmann exploration policy satisfies the above two conditions. Then we provide the result from Singh \textit{et al.}.

\begin{lemma}(\cite{singh2000convergence})
\label{sarsalemma}
Let $M$ be a finite MDP and $\pi$ is a GLIE policy. If the step-size sequence $\{\alpha_t(s_t,a_t)\}$ such that 
\begin{equation*}
\sum_t \alpha_t(s_t,a_t)=\infty \qquad \sum_t \alpha_t(s_t,a_t)^2<\infty,
\end{equation*}
Then, the Sarsa Algorithm converges with probability 1.
\end{lemma}

\begin{proof}
\cite{singh2000convergence} prove a similar convergence result under a weaker assumption, they assume that $Var(r(s,a))< \infty$ . In this paper, we assume that $r(s,a)$ is bounded, that is $|r(s,a)|<\infty$ for all $(s,a)$ pairs, which implies $Var(r(s,a))< \infty$. 
\end{proof}

\begin{theorem}
Given an advice model $(T,S,\pi_d)$, the student $S$ adopts the Sarsa Algorithm and conditions in Lemma \ref{sarsalemma} all hold, convergence of Sarsa still holds in the advice model setting
\end{theorem}

\begin{proof}
Notice that the GLIE policy guarantee that each state-action pair is visited infinitely many times. And there is finite advice in our advice model. Therefore, the assumptions still hold in advice model setting. Apply Lemma \ref{sarsalemma}, the convergence result follows. 
\end{proof}

\begin{remark}
On one hand, the key for the convergence results is that each state-action pair is visited infinitely often. For an advice model, the finite budget does not invalidate the infinite visit assumption. Therefore, the results follows from previous convergence results hold. On the other hand, the infinite visit assumption is a sufficient condition for the convergence result --- if the assumption does not hold, the convergence may still hold. Moreover, the algorithms converge even if the budget is infinite as long as the student is still able to visit all state-action pairs infinitely many times. 
\end{remark}

\subsection{Linear Function Approximation}
In the previous subsection, we discuss some results regarding tabular representation learning algorithms that require an MDP with finite states and actions at each state. However, infinite or large state-action space in practice is very important since they can characterize many realistic scenarios. 

The convergence of Q-learning and Sarsa with linear approximation in standard setting has been proved \cite{melo2008analysis}, provided the relevant assumptions hold.
%They provide conditions in which Q-learning and Sarsa with linear approximation converges.
Our approach is inspired by this work, which assumes that the algorithm (Q-learning or Sarsa, with linear approximation) holds under the convergence conditions in \cite{melo2008analysis}. %\MET{Can we enumerate the assumptions here, or are we out of space?} 
We then apply the convergence theorems to the action advice model $(T,S,\pi_d)$, and the results follow.

We need to define some notations for simplifying our proofs. Given an MDP $M=(S,A,P,R,\gamma)$ with a compact state set $S$ and a fixed policy $\pi$, $\mathcal{M}=(S,P_{\pi})$ is the Markov chain induced by policy $\pi$.  Assume that the chain $\mathcal{M}$ is uniformly ergodic with invariant probability measure $\mu_S$ over $S$ and the policy $\pi$ satisfies $\pi(s,a)>0$ for all $a\in A$ and all $s\in S$ with non-zero $\mu_S$ measure\footnote{This condition is able to be interpreted as the continuous counterpart of ''infinite visit'' in finite action-state space scenario.}. Let $\mu_{\pi}$ be the probability measure for all Borel-measurable set $U \subset S$ and for all action $a \in A$, 
$$\mu_{\pi}(U\times\{a\})=\int_U \pi(s,a)\mu_S(ds).$$

Suppose that $\{\phi_i\}_{i=0}^d$ is a set of bounded, linearly independent features, we define matrix $\Sigma_{\pi}$ as 
$$\Sigma_{\pi}=\mathbb{E}[\pmb\phi^\intercal(s,a)\pmb\phi(s,a)]=\int_{S\times A}\pmb\phi^\intercal(s,a)\pmb\phi(s,a)d\mu_{\pi}$$
Notice that $\Sigma_{\pi}$ is independent of the initial probability distribution due to uniform ergodicity.

For a fixed $\pmb\theta\in \mathbb{R}^d$, $d>1$ and a fixed state $s \in S$, define the set of optimal actions in state $s$ as 
$$A_{\pmb\theta,s}=\left\{a^\star\in A \middle| \pmb\theta^\intercal\pmb\phi(s,a^\star)=\max_{a\in A}\pmb\theta^\intercal\pmb\phi(s,a)\right\}.$$ 
A policy $\pi$ is greedy w.r.t. $\pmb\theta$ which assigns positive probability only to actions in $A_{\pmb\theta,s}$. We define $\pmb\theta$-dependent matrix $$\Sigma_{\pi}^\star(\pmb\theta)=\mathbb{E}_{\pi}\left[ \pmb\phi^\intercal(s,a_{\pmb\theta,s})\pmb\phi(s,a_{\pmb\theta,s})\right],$$ where $a_{\pmb\theta,s}$ is a random action determined by policy $\pi$ at state $s$ in set $A_{\pmb\theta,s}$. Notice that the difference between $\Sigma_{\pi}$ and $\Sigma_{\pi}^\star(\pmb\theta)$ is that the actions are taken according to $\pi$ in $\Sigma_{\pi}$ while in $\Sigma_{\pi}^\star(\pmb\theta)$ they are taken greedily w.r.t. a fixed $\pmb\theta$, that is, actions in $A_{\pmb\theta,s}$. 

We will show that Q-learning with linear function approximation still converges in the advice model setting at first. We introduce following lemma: 

\begin{lemma}(\cite{melo2008analysis} Theorem 1)
\label{q-learninglinaerlemma}
Let $M$, $\pi$ and $\{\phi_i\}_{i=0}^d$ be defined as above. if, for all $\pmb\theta$, $\pmb\Sigma_{\pi} >\gamma^2 \pmb\Sigma_{\pi}^\star(\pmb\theta)$ and the step-size sequence $\{\alpha_t(s_t,a_t)\}$ such that 
\begin{equation*}
\sum_t \alpha_t(s_t,a_t)=\infty \qquad \sum_t \alpha_t(s_t,a_t)^2<\infty,
\end{equation*}
then the Algorithm Q-learning with linear approximation converges with probability 1
\end{lemma}

\begin{theorem}
Given an advice model $(T,S,\pi_d)$, if the Markov chain which is induced by $\pi_d$ is also \textbf{uniformly ergodic} and the student $S$ adopts the Q-learning with linear approximation and conditions in Lemma \ref{q-learninglinaerlemma} all hold, the convergence of Q-learning with linear approximation still hod in the advice model setting.
\end{theorem}

\begin{proof}
Apply Lemma \ref{q-learninglinaerlemma},  the convergence result still hold in the advice model setting.
\end{proof}

Next, we will analyze the convergence of Sarsa with linear approximation in the advice model. Sarsa is an on-policy algorithm, we need some different assumptions. A policy $\pi$ is $\epsilon$-greedy with respect to a Q-value function $Q$ for a fixed $\pmb\theta$, if it chooses a random action with probability $\epsilon>0$ and a greedy action $a \in A_{\pmb\theta,s}$ for all state $s \in S$. 
A $\pmb\theta$-dependent policy $\pi_{\pmb\theta}$ satisfies $\pi_{\pmb\theta}(s,a)>0$  for all $\pmb\theta$.
Now we consider a policy $\pi_{\pmb\theta_t}$ is $\epsilon$-greedy with respect to $\pmb\theta_t^\intercal\pmb\Phi(s,a)$ at each time step $t$ and Lipshitz continuous with respect to $\pmb\theta$, where $K$ denotes the Lipshitz constant (refers to a specific metric)\footnote{Given two metric spaces $(X,d_x)$ and $(Y,d_y)$, where $d_X$ and $d_Y$ denotes metric on set $X$ and $Y$, respectively. A function $f:X \to Y$ is called \textbf{Lipshitz continuous}, if there exists a real constant $K\geq 0$ such that for all $x_1,x_2 \in X$, $$d_Y(f(x_1),f(x_2))\leq Kd_X(x_1,x_2),$$ where the constant $K$ is called \textbf{Lipshitz constant}.}. Moreover, we assume that induced Markov chain $\mathcal{M}=(S,P_{\pmb\theta})$ is uniformly ergodic.

\begin{lemma}(\cite{melo2008analysis} Theorem 2)
\label{sarsalinearlemma}
Let $M$, $\pi_{\theta_t}$ and $\{\phi_i\}_{i=0}^d$ be defined as above. Let $K$ be the Lipshcitz constant of the learning policy $\pi_{\theta}$ w.r.t. $\theta$. If the step-size sequence $\{\alpha_t(s_t,a_t)\}$ such that 
\begin{equation*}
\sum_t \alpha_t(s_t,a_t)=\infty \qquad \sum_t \alpha_t(s_t,a_t)^2<\infty,
\end{equation*}
Then, there is $K_0>0$ such that, if $K<K_0$, the Sarsa with linear approximation converges with probability 1.
\end{lemma}

\begin{theorem}
Given an advice model $(T,S,\pi_d)$, if $\pi_d$ is $\pmb\theta$-dependent and $\epsilon$-greedy w.r.t. a fixed $\pmb\theta_t$ at each time step $t$.  The student $S$ adopts the Sarsa with linear approximation and conditions in Lemma \ref{sarsalinearlemma} all hold, Sarsa with linear approximation still converges with probability 1.
\end{theorem}

\begin{proof}
Apply Lemma \ref{sarsalinearlemma},  the convergence result still hold in the advice model setting.
\end{proof}

\begin{remark}
Notice that we assume the budget of the teacher is finite which implies that any finite policies do not affect the convergence results as long as the conditions in Lemma \ref{q-learninglemma}, \ref{sarsalemma}, \ref{q-learninglinaerlemma} and \ref{sarsalinearlemma} still hold. Therefore, the student will eventually converge even if the teacher is sub-optimal.
\end{remark}

%\subsection{Convergence Results}
%The convergence of Q-learning and Sarsa with tabular representation or linear function approximation in the advice model relies on some assumptions. We summaries as following:
%
%\begin{itemize}
%\item Q-learning with tabular representation: there are a lot of works about the convergence of Q-learning with tabular representation \cite{watkins1992q,jaakkola1994convergence,tsitsiklis1994asynchronous,mohri2012foundations}. They usually require the Q-learning algorithm is capable of accessing state-action pair infinitely. Since the advice model does not violate this assumption, the result follows.
%\item Sarsa with tabular representation: Sarsa is a on-policy algorithm which requires a learning policy to update the Q values. \cite{singh2000convergence} prove that Sarsa with GLIE, greedy in the limit with infinite exploration, policy converges. The advice model is compatible with the GLIE assumption, the result follows.
%\item Q-learning and Sarsa with linear function approximation: the convergence of Q-learning and Sarsa with linear approximation in standard setting has been proved \cite{melo2008analysis}, providing the relevant assumptions hold. Therefore, we assume that those assumptions still hold under the advice model, the convergence results follow.
%\end{itemize}

\subsection{Asymptotic Performance}
Next, we will investigate the asymptotic behavior in the advice model. Most of convergence results rely on infinite experience, which is not suitable in practice --- we first redefine the concept of convergence.

\begin{definition}[Convergence in Algorithm Design]
\label{convergence}
If an algorithm $\mathfrak{A}$ converges, then there exits a $N \in \mathbb{N}$, for all $t\geq N$ such that 
\begin{equation*}
||Q_{t+1}-Q_t||_{\infty} \leq \epsilon,
\end{equation*}
where $\epsilon$ is very small constant.
\end{definition}

\begin{theorem}
\label{asymptotictheorem}
If an algorithm $\mathfrak{A}$ converges in terms of Definition \ref{convergence}, then finite advice cannot improve the asymptotic performance of algorithms $\mathfrak{A}$.
\end{theorem}

\begin{proof}
If an algorithm $\mathfrak{A}$ converges, then there is a $N \in \mathbb{N}$  for all $t\geq N$ such that 
\begin{equation*}
||Q_{t+1}-Q_t||_{\infty} \leq \epsilon,
\end{equation*}
where $\epsilon$ is very small constant.
%\MET{I'm confused here. Was there another sentence or two of text here before? Right now, I don't see how just because an algorithm converges, it's guaranteed to converge to the same performance with or without a finite amount of advice.}
Therefore, even if the advice is sub-optimal the student will always find the optimal action according to its own Q-value after $N$ updates, that is, finite advice can not affect the asymptotic performance in the sense of infinite horizon. The asymptotic performance is determined by the algorithms that the student uses, not the advice provided by a teacher. 
\end{proof}

\begin{remark}
Theorem \ref{asymptotictheorem} indicates the limitation of the advice model. Generally, there are two intuitive methods to improve the performance of student in the advice model: (1) higher amounts of advice, or (2) redistribution of the advice(e.g., delay the advice for when it is most useful). Our theorem points out that, with a finite budget for advice,
%\MET{with a finite amount of advice}, 
%if the budget exceeds a certain threshold, 
the asymptotic performance is still determined by the algorithm that the student adopts as long as the algorithm converges. Furthermore, advice delay is limited also due to the convergence of the algorithm that the student uses.
\end{remark}

\section{Experimental Domain and Results}
\iffalse
\begin{figure}[t]
\centering
\includegraphics[width=0.5\linewidth]{fig/1.pdf}
\caption{Pac-Man}
\label{pacmanteachhuman}
\end{figure}
\fi
In this section, we introduce the experimental results in two domains. The goal of experiments is to provide \textbf{experimental support for convergence proofs} from the previous section, as well as to justify that \textbf{action advice improves learning}. The first domain is a simple linear chain of states: Linear Chain. The second is Pac-Man, a well-known arcade game. We will apply Q-learning with tabular learning to the Linear Chain and Sarsa with linear function approximation to Pac-Man. 

%\begin{figure*}
%\centering
%\mbox{\epsfxsize=3.3in \epsfbox{Linear_MDP.eps}}
%\qquad
%\mbox{\epsfxsize=3.3in \epsfbox{pacman.eps}}
%\caption{Top: Q-learning students in Linear Chain MDP. Bottom: Sarsa students in Pac-Man domain.}
%\label{linear_mdp}
%\label{pacman}
%\end{figure*}

\begin{figure}
	\centering
	\includegraphics[width=0.8\linewidth]{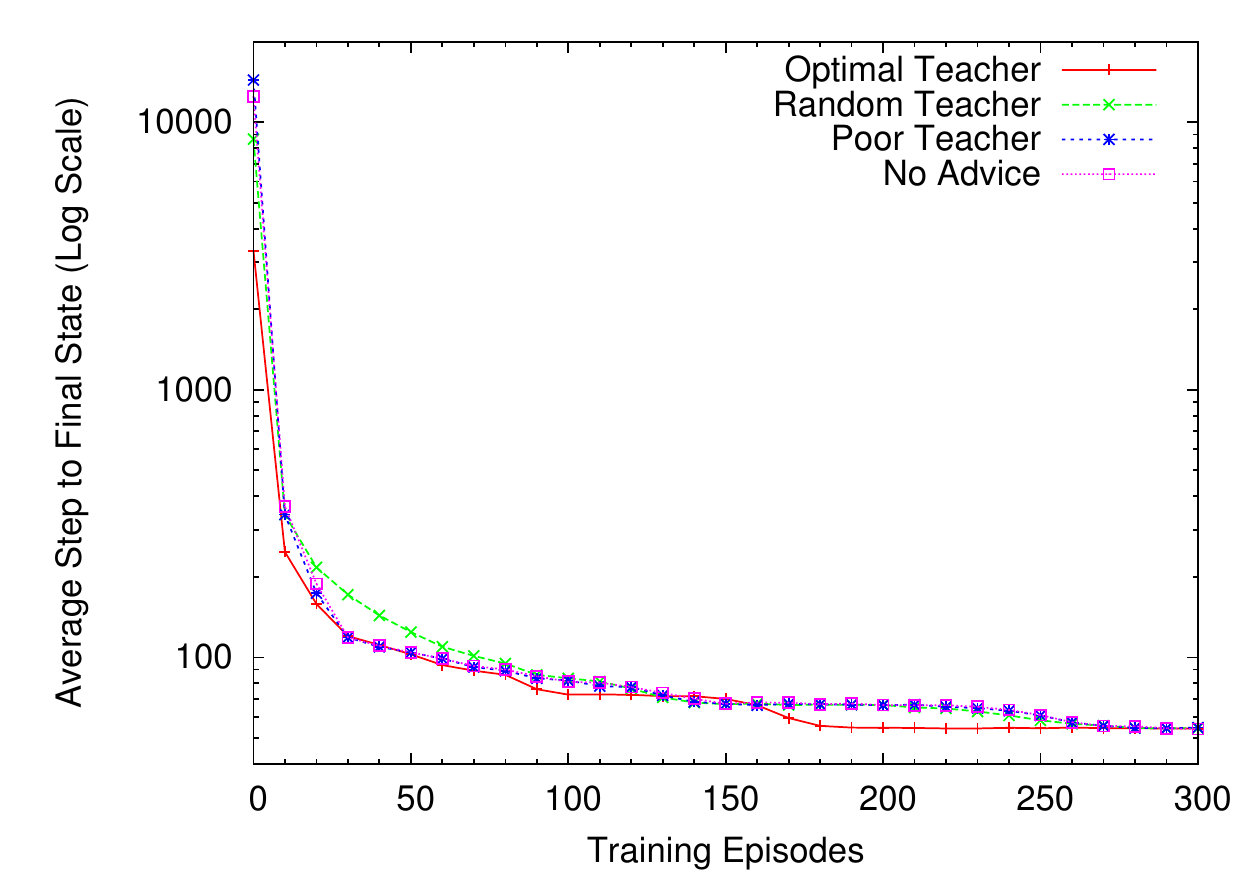}
	\includegraphics[width=0.8\linewidth]{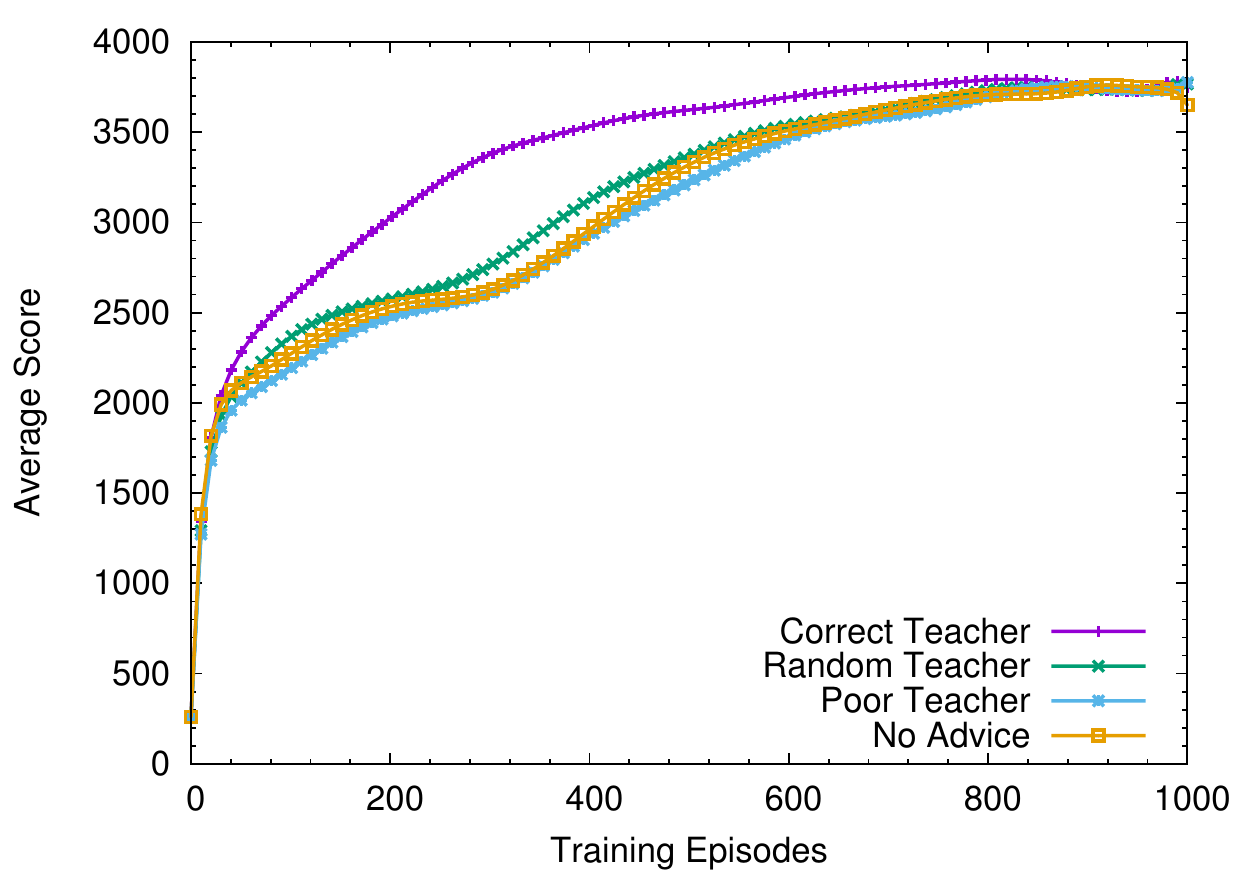}
	\caption{Top: Q-learning students in Linear Chain MDP. Bottom: Sarsa students in Pac-Man domain.}
	\label{linear_mdp}
	\label{pacman}
\end{figure}

\subsection{Linear Chain MDP}
\begin{figure}[t]
\centering
\begin{tikzpicture}[start chain=double]
\node[state, on chain]                 (1) {0};
\node[state, on chain]                 (2) {2};
\node[on chain]                   (2-g) {$\dots$};
\node[state, on chain]                 (48) {48};
\node[state, on chain]                 (49) {49};

% The \draw path is like the one above.

\draw[latex'-latex',double]

     (1)   edge                    node {}   (2)
            %edge [bend left]   node {$1-a$}   (0)
     (2)  edge                    node {}   (2-g)
     		%edge [bend left]   node {$1-a$}   (0)
     (2-g)   edge                node {}   (48)
     
     (48) edge                     node{} (49);
\end{tikzpicture}
\caption{Linear Chain MDP with $50$ states. State $0$ is the start state and  state $49$ is the goal state.}
\label{linearchainmdp}
\end{figure}
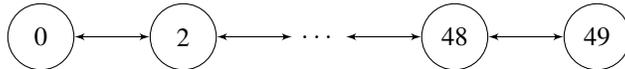

The first experimental domain is the Linear Chain MDP~\cite{lagoudakis2003least}. In this domain, we adopt Q-learning with the tabular representation to store the Q-values due to the simplicity. See Figure \ref{linearchainmdp} for details. 

In this paper, the MDP has $50$ states and two actions for each state: left and right.  state $0$ is the start state and state $49$ is the final state in which the episode is terminated. The agent will receive $-1$ reward per step in non-terminated states and $0$ in goal state. 

To smooth the variance in student performance, we average $300$ independent trials of student learning. Each Linear Chain teacher is given an advice budget of $n=1000$. The reinforcement learning parameters of the students are $\epsilon=0.1$, $\alpha=0.9$ and $\gamma=0.8$.

We use four experimental setting to demonstrate the convergence results:
\begin{itemize}
\item Optimal Teacher: The teacher will always give the optimal action in each state, i.e., move right.
\item Random Teacher: The teacher will give action advice, 50\% move left and 50\% move right.
\item Poor Teacher: The poor teacher gives the worst action, e.g., move left.
\item No Advice: There is no advice, equivalent to normal reinforcement learning.
\end{itemize}

Figure \ref{pacman} (top) shows the results of these experiments (note the log scale on the y-axis). All settings converge after $280$ episodes training despite different teacher performance. 

To compare methods, we calculate the area under each learning curve. We apply one-way ANOVA to test the difference between all settings and the result shows that the $p<2\times 10^{-16}$, indicating that we should reject the null hypothesis that ``all test groups have same means.'' Therefore, all experimental settings are statistically different, where the optimal teacher outperforms the random teacher, which outperforms no advice, which outperforms the poor teacher. Also, we provide the final reward, standard deviation of final reward, total reward and standard deviation of final reward on Table \ref{linearmdptabe}.

\begin{table}
\centering
\scalebox{0.85}{
\begin{tabular}{|c|c|c|c|c|}
\hline 
Group & FR & FR STD & TR & TR STD \\ 
\hline 
Optimal Teacher & $-53.99$ & $3.30$ & $-29007.01$ & $1384.31$ \\ 
\hline 
Random Teacher & $-54.56$ & $3.59$ & $-41670.98$ & $2398.87$ \\ 
\hline 
Poor Teacher & $-54.28$ & $3.58$ & $-43964.97$ & $2394.78$ \\ 
\hline 
No Advice & $-54.13$ & $3.221$ & $-42355.24$ & $2660.75$ \\ 
\hline 
\end{tabular} 
}
\caption{FR is the final reward of the last episode, FR STD is final reward's standard deviation, TR is the total reward accumulated reward in all episodes, and TR STD is the standard deviation of total reward.}
%%With bold typeface are highlighted the best achieved time
%%in 6 vs.\ 5 and the best total time. }
\label{linearmdptabe}
\end{table}

\subsection{Pacman}

Pac-Man is a famous 1980s arcade game in which the player navigates a maze, trying to earn points by touching edible items and trying to avoid being caught by the four ghosts.
We use a JAVA implementation of the game provided by the Ms.\ Pac-Man vs.\ Ghosts League~\cite{rohlfshagen2011ms}. This domain is discrete but has a very large state space due to different position combination of player and all ghosts --- linear function approximation is used to represent state.
Student agents learn the task using Sarsa and a state representation defined by $7$ features that count objects at a range of distances, as used (and defined) in~\cite{torrey2013teaching}. 

To smooth the natural variance in student performance, each learning curve averages $30$ independent trials of student learning. While training, an agent pauses every few episodes to perform at least $30$ evaluation episodes and record its average performance --- graphs show the performance of students when they are 1) not learning and 2) not receiving advice.

Each Pac-Man teacher is given an advice budget of $n=1000$, which is half the number of the step limit in a single episode. The reinforcement learning parameters of the students are $\epsilon=0.05$, $\alpha=0.001$ and $\gamma=0.999$.

To demonstrate that finite advice can not affect the convergence of students, we adopt different experimental settings:
\begin{itemize}
\item Correct Teacher: Provide the (near-)optimal action when it observes the student is about to execute a sub-optimal action.
\item Random Teacher: Provide random action suggestion from the set of legal moves.
\item Poor Teacher: Advise the student to take the action with the lowest Q-value whenever the student is about to execute a sub-optimal action.
\item No Advice: There is no advice, equivalent to normal reinforcement learning.
\end{itemize}

See the experimental results in Figure \ref{pacman} (bottom).  All settings converges after $900$ episodes training despite different teacher performance. 
As before, a one-way ANOVA is used to test the total reward accumulated by the four different teaching conditions. $p < 4.6\times 10^{-13}$, showing that all experimental settings are statistically different, and that the correct teacher was better than no advice, which was better than the random teacher, which was better than the poor teacher. Also, we provide rewards on Table \ref{pacmantabel}.

\begin{table}
\centering
\scalebox{0.85}{
\begin{tabular}{|c|c|c|c|c|}
\hline 
Group & FR & FR STD & TR & TR STD \\ 
\hline 
Correct Teacher & $3746.75$ & $192.18$ & $341790.99$ & $5936.23$ \\ 
\hline 
Random Teacher & $3649.78$ & $167.86$ & $313151.06$ & $4634.88$ \\ 
\hline 
Poor Teacher & $3775.13$ & $148.34$ & $307926.03$ & $7708.45$ \\ 
\hline 
No Advice & $3766.58$ & $132.41$ & $318072.70$ & $7660.44$ \\ 
\hline 
\end{tabular} 
}
\caption{FR, FR STD, TR and TR STD are same as those in Table \ref{linearmdptabe}.}
%\caption{Final Reward(FR)is the last episode, FR STD is final reward standard deviation, Total Reward(TR) is accumulated reward for all episodes and TR STD is the standard deviation of total reward}
%With bold typeface are highlighted the best achieved time
%in 6 vs.\ 5 and the best total time. }
\label{pacmantabel}
\end{table}

\section{Related Work}
This section briefly outline related work in transfer learning in reinforcement domains, online transfer learning in supervised learning, and algorithmic teaching.

Transfer learning in reinforcement domain has been studies recently \cite{taylor2009transfer,transferalessan}. Lazaric introduces a transfer learning framework which inspires us to develop the online transfer learning framework and classifies transfer learning in reinforcement domain into three categories: instance transfer, representation transfer and parameter transfer~\cite{transferalessan}. The action advice model is a method of instance transfer due to explicit action advice (i.e., sample transfer). Lazaric proposed an instance-transfer method which selectively transfers samples on the basis of the similarity between source and target tasks~\cite{lazaric2008transfer}.

\citeauthor{azar2013regret} (\citeyear{azar2013regret}) introduced a model that takes the teacher/advice model as input and a learning reinforcement learning algorithm is able to query the input advice policy as it is necessary. However, their model does not consider the learning reinforcement learning algorithm behavior, which we believe is important in online reinforcement learning.  

Zhao and Hoi propose an online transfer learning framework in supervised learning \cite{zhao2010otl}, aiming to transfer useful knowledge from some source domain to an online learning task on a target domain. They introduce a framework to solve transfer in two different settings. The first is that source tasks share the same domain as target tasks and the second is that the source domain and target domain are different domain.

Finally, a branch in computational learning theory called algorithmic teaching tries to understand teaching in theoretical ways \cite{balbach2009recent}. In algorithmic learning theory, the teacher usually determines a example sequence and teach the sequence to the learner. There are a lot of algorithmic teaching models such as teaching dimension \cite{goldman1995complexity} and teaching learners with restricted mind changes~\cite{balbach2005teaching}. However, those models still concentrate on supervised learning. \cite{cakmak2012algorithmic} developed a teaching method which is based on algorithm teaching, but their work focuses on one-time optimal teaching sequence computing, which lacks the online setting.

\section{Discussion}
\label{sec:discussion}
This paper proposes an online transfer learning framework. It then characterizes two existing works addressing teaching in reinforcement learning. A theoretical analysis of one of the methods, where teachers provide action advice,  lead us to the following conclusions. First,
Q-learning and Sarsa converge to the optimal Q-value%\MET{To the optimal performing policy?} 
when there is a finite amount of advice. Second, with linear function approximation, Q-learning and Sarsa converge to the optimal Q-value %\MET{to the optimal performing policy?}
, assuming normal assumptions hold. Third,there is a limit of the advice model: teacher advice can not affect the asymptotic performance of any algorithms that converge. Fourth, our results are empirically justified in the Linear Chain MDP and in Pac-Man.

In the future, sample complexity and regret analysis for the advice model will be investigated, now that the convergence results have been established. Additional models under the online transfer framework will be developed, which will not only focus on interaction between machines, but also consider interaction between machines and humans (e.g., learning from demonstration~\cite{argall2009survey}). Finally, we will consider other reinforcement learning algorithms such as R-Max and study the theoretical properties of those algorithms in the presence of the advice model.

\section{Acknowledgments}
This research has taken place in the Intelligent Robot Learning (IRL)
Lab, Washington State University. IRL research is support in part by
grants from AFRL FA8750-14-1-0069, AFRL FA8750-14-1-0070, NSF
IIS-1149917, NSF IIS-1319412, and USDA 2014-67021-22174.
s supported in part by NSF IIS-1149917.
%Thank Alessandro Lazaric, Michael Littman, and Haitham Bou Ammar

%\MET{The cakmak/lopes citation is incorrect}
%\MET{Please make sure all the citations look the same. For instance, some have page numbers (Zhao and Hoi) and some don't (Cakmak, Lopes). Also, Sometimes there's the full conference name (Lazaric, Restelli, and Bonarini) and sometimes just the abbreviation (cakmak, Lopes).}

\bibliographystyle{cell}
\bibliography{General}

\end{document}